%% file: main.tex
\documentclass[10pt,twocolumn,letterpaper]{article}

\usepackage[pagenumbers]{cvpr} 

\input{preamble}

\definecolor{cvprblue}{rgb}{0.21,0.49,0.74}
\usepackage[pagebackref,breaklinks,colorlinks,citecolor=cvprblue]{hyperref}

\def\paperID{*****} 
\def\confName{CVPR}
\def\confYear{2024}

\title{Rethinking Overlooked Aspects in Vision-Language Models}

\author{
  \hspace{-0.25cm}\textbf{Yuan Liu, Le Tian, Xiao Zhou, Jie Zhou} \\
  \hspace{-0.25cm}Pattern Recognition Center, WeChat AI, Tencent Inc, China \\
  \hspace{-0.25cm}\tt\small\{bensenliu, letian, chappyzhou, withtomzhou\}@tencent.com \\
  }

\begin{document}
\maketitle
\input{sec/0_abstract}  
\input{sec/1_intro}

\input{sec/2_related}

\input{sec/3_method}
\input{sec/4-exp}

\input{sec/5-discussion}

\input{sec/6-conclusion}

{
    \small
    \bibliographystyle{ieeenat_fullname}
    \bibliography{main}
}


\end{document}

%% file: preamble.tex
\usepackage[dvipsnames]{xcolor}
\usepackage{color, colortbl}
\usepackage{arydshln}
\usepackage{multirow}
\usepackage{tabularx}

\definecolor{COLOR_MEAN}{HTML}{f0f0f0}
\definecolor{PURPLE}{HTML}{8983bf}
\definecolor{LIGHT_BLUE}{HTML}{cce4fe}
\definecolor{LIGHT_RED}{HTML}{f1b9b8}
\definecolor{LIGHT_YELLOW}{HTML}{f1f58a}
\definecolor{green}{RGB}{84,179,69}
\newcommand{\more}[1]{\footnotesize {\textcolor[RGB]{57,181,74}{#1}}}
\newcommand{\less}[1]{\footnotesize {\textcolor[RGB]{255,0,0}{#1}}}

%% file: sec/0_abstract.tex
\begin{abstract}
In recent years, large vision-language models (LVLMs), such as GPT4-V, have advanced significantly, primarily due to the transformative impact of large language models. Among these LVLMs, LLaVA stands out as a widely adopted model, offering several key advantages: (i) \textit{Simplicity}—LLaVA consists of three main components: a vision encoder, a lightweight adapter (e.g., MLP), and a large language model (LLM). This modular architecture facilitates the integration of state-of-the-art (SOTA) models into any component to boost performance. (ii) \textit{Efficiency}—With a modest pre-training dataset of 558k images and 665k instances of supervised fine-tuning data, LLaVA-1.5-13B achieves impressive results across numerous benchmarks. Subsequent works have begun to (i) incorporate substantially more data during pre-training, and (ii) utilize a more diverse and larger instruction tuning dataset. In this report, we aim to investigate two important, yet previously overlooked, aspects: (i) \textbf{the efficiency of data during pre-training}—whether the model's performance consistently improves with the addition of more pre-training data; (ii) \textbf{how to choose instruction tuning datasets}—the effectiveness of the SFT datasets used in existing works and the methodology for selecting the most impactful ones. Through a meticulous and comprehensive examination, we have discovered that a naive increase in the size of the pre-training dataset does not effectively enhance the performance of Vision-Language Models (VLM). In fact, it may even lead to a degradation in performance. Regarding the SFT data, we have developed an effective pipeline to identify the most efficient SFT dataset. Our study reveals that not all SFT data employed in existing works are necessary and can be optimized for enhanced performance. With the findings of this report, we hope to encourage future research to focus more on the data used during pre-training and supervised fine-tuning to further push the boundaries of vision-language models.

\end{abstract}

%% file: sec/1_intro.tex
\section{Introduction}
\label{sec:intro}
Large language models (LLMs) have achieved significant progress in recent years. Some models, such as GPT-4~\cite{achiam2023gpt} and Claude3, have reached or even surpassed human performance in various aspects. Compared to LLMs, large vision-language models (LVLMs) can solve much more complex problems that the original text-only LLMs could not, such as image understanding and question-answering. With the rapid development of LLMs, LVLMs have also demonstrated remarkable achievements. Models like the recent GPT-4o, Qwen-VL-Max, and Step-1V show promising capabilities in solving increasingly complex image-related problems, including geometry matching and optical character recognition. However, all these models are proprietary, and the details behind them are not publicly known. Despite the existing gap between open-source models and these proprietary models, recent progress~\cite{chen2024far, liu2024llavanext} has been made, which is gradually narrowing this divide.

\input{figures/teaser}

We find that the latest advancements in visual language models are largely driven by data, including pre-training data and instruction tuning (SFT) data. For instance, models like InternVL-1.5\cite{chen2024far}, Qwen-VL-Max\cite{bai2023qwen}, and DeepSeek-VL\cite{lu2024deepseek} utilize web-scale pre-training datasets such as Laion-5B\cite{schuhmann2022laionb} and COYO\cite{kakaobrain2022coyo-700m}, enabling them to reach a pre-training data volume of 1B. Simultaneously, compared to previous works, their instruction tuning datasets are not only larger in scale but also richer in diversity. For example, InternVL-1.5 divides the sft dataset into 11 subclasses and collects corresponding open-source datasets for each subclass, a practice also adopted by DeepSeek-VL. For \textbf{pre-training datasets}, there exists a scaling law in the LLM field\cite{hoffmann2022training}, which suggests that as the model size increases and the pre-training dataset size is concurrently expanded, the model's performance will also increase synchronously. However, no work in the vision-language model field has yet conducted a comprehensive experiment of this nature. Therefore, we are curious: 1) For the same model, if we continuously increase the amount of pre-training data using existing open-source multimodal datasets, will the model's performance also grow synchronously? 2) When we enlarge the model size and concurrently increase the pre-training data volume, will visual language models also demonstrate a scaling law similar to that in LLM? Regarding the \textbf{SFT datasets}, for a long time, everyone has been conducting instruction-following training based on the dataset proposed in LLaVA-1.5\cite{liu2023improved}. Compared to the data in LLaVA-1.5, the latest works have introduced datasets with more categories and larger quantities. However, given the varying quality of these newly introduced datasets and the significant overlap between different datasets, it raises the question of whether these datasets all play a key role in enhancing the model's general capabilities.

LLaVA-1.5 is a very simple and efficient model, mainly composed of three parts: a vision encoder, a vision-language adapter, and a large language model. Upon its introduction, it receives significant attention and has been used and improved upon in many subsequent works. We designed several experiments based on LLaVA-1.5, the simplest model, hoping to reveal answers to the above questions. Firstly, regarding the pre-training dataset, we extracted seven sets of data from LAION-5B-en, with sizes ranging from 1M to 100M, and trained the same model on these datasets. Simultaneously, to observe whether the model's performance steadily improves with the increase in model size and data volume, we select Vicuna-7B/13B, Qwen1.5-Chat-7B/14B, and Yi-Chat-6B/34B to study this phenomenon. As for the SFT dataset, we used the dataset in LLaVA-1.5 as the base version. Referring to the taxonomy of the SFT dataset in InternVL-1.5\cite{chen2024far}, we proposed a method called \textbf{Individual Select}, which selects the most effective dataset in each category on a single dataset granularity. Through a large number of experiments, we have found that:
\begin{enumerate}[label={\bf {{(\arabic*)}}},leftmargin=*,topsep=0.5ex,itemsep=-0.5ex,partopsep=0.75ex,parsep=0.75ex,partopsep=0pt,wide, labelwidth=0pt,labelindent=0pt]
    \item When we naively use existing datasets to increase the amount of pre-training data for the model, the performance of the model does not improve, and may even cause a decline in the model's performance.
    \item For the same type of models, such as Vicuna-7B and Vicuna-13B, increasing the model size while also increasing the amount of pre-training data in parallel, the ultimate improvement in model performance is largely due to the use of a larger model. 
    \item The SFT dataset used in the latest work has a lot of redundancy, and there is still a large exploration space. 
\end{enumerate}

In summary, the aim of this report is not to propose a state-of-the-art (SoTA) model, but rather to explore crucial yet previously overlooked facets of vision-language development and research. We anticipate that the findings from this report will offer valuable insights and guidance for the future advancement of vision-language models.

%% file: figures/teaser.tex
\begin{figure}[t!]
\centering
\includegraphics[width=\linewidth]{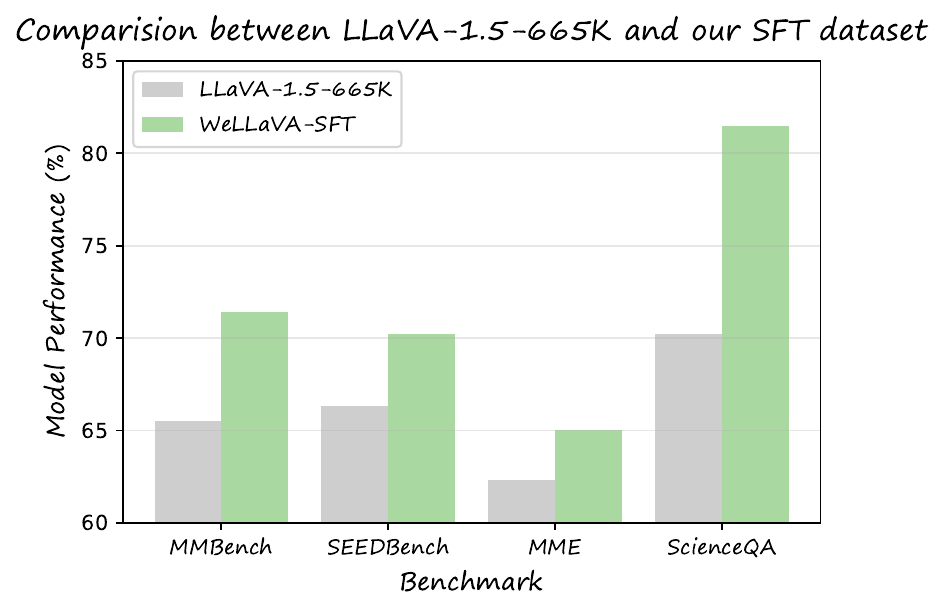}
\vspace{-2em}
\caption{\textbf{Performance of LLaVA-1.5 using LLaVA-1.5-665K and our SFT dataset.} We have developed a strategy, termed \textbf{Individual Select}, which is designed to select the most effective datasets from a plethora of publicly available SFT datasets. The LLaVA-1.5 model, fine-tuned with Vicuna-7B\cite{vicuna2023} on the final composition of SFT datasets that we have obtained, yields substantial improvements compared to the baseline. The original MME scores are mapped to a range of 0 to 100.}
\label{fig:teaser}
\end{figure}

%% file: sec/2_related.tex
\section{Related Works}
\label{sec:related}

\paragraph{Pre-training} Pre-training is a prevalent technique across various domains, such as computer vision and natural language processing. The primary objective of pre-training is to equip a randomly initialized model with the capability to accomplish general tasks. For instance, in computer vision, tasks like classification and instance segmentation often benefit from pre-training the model using methods like masked image modeling\cite{he2022masked, liu2023improving, liu2023pixmim}. Moreover, the recent advancements in large language models can be attributed to extensive pre-training on vast datasets. In the context of vision-language models (VLMs), they typically comprise three components: a vision encoder, an adapter, and a language model. Given that the vision encoder and language model are trained separately, they exhibit different feature distributions. Therefore, the primary focus of a vision-language model is on vision-language alignment. During this alignment process, some studies\cite{chen2024allava, liu2023improved, chen2023sharegpt4v} opt to freeze both the vision encoder and language model, training only the lightweight adapter with a smaller dataset. Conversely, others choose to freeze only the language model, training both the vision encoder and adapter with a larger dataset. However, there exists no agreement about how much data should be used during pre-training.

\paragraph{Instruction Tuning} Instruction tuning is a crucial technique in both large language models (LLMs) and vision-language models (VLMs), enabling the model to accurately follow human instructions. However, research on instruction tuning in VLMs significantly lags behind that in LLMs. LLaVA\cite{das2017visual}, a pioneering work in this field, suggests using text-only GPT4 to generate a large number of instruction tuning datasets. The instruction tuning dataset in LLaVA primarily comprises three categories: conversation, complex reasoning, and detailed description. Subsequently, InstructBLIP\cite{dai2024instructblip} proposed the use of a vast amount of academic datasets as instruction tuning datasets and designed specific prompts for each dataset. Models fine-tuned on these datasets demonstrated impressive performance at the time. Following this, a series of recent works\cite{internlmxcomposer2, liu2023improved, chen2023internvl, chen2024far} began to incorporate datasets from various sources, such as those generated by GPT-4V and academic datasets, further elevating the performance of vision-language models to unprecedented levels. However, it is unfortunate that, to date, no study has provided insights into the type of SFT dataset that is most efficient, or whether we need all the datasets to fine-tune our model.

%% file: sec/3_method.tex
\section{Overlooked Aspects}
\label{sec:method}

\subsection{Creating a Stable Baseline}

\paragraph{Benchmarks Selection.} We anticipate that the model will excel in a variety of general tasks, rather than concentrating on a single specific task. Consequently, we expect a benchmark should be able to evaluate different aspects of a model comprehensively. Recently proposed benchmarks such as MMBench\cite{liu2023mmbench}, MME\cite{Fu2023MMEAC}, and SEED-Bench are designed to provide a thorough evaluation of a model's performance. In comparison to earlier benchmarks\cite{goyal2017making, hudson2019gqa}, these cover a wider range of ability dimensions, offering a more comprehensive insight into the capabilities of the evaluated model. Therefore, we choose these benchmarks as our guidelines for setting selection during our exploration. Furthermore, considering the frequent appearance of scientific problems in our daily lives, we have also included the metric from ScienceQA\cite{lu2022learn} in our guidelines. All the subsequent results are obtained using the evaluation toolkit, VLMEvalKit\footnote{\href{https://github.com/open-compass/VLMEvalKit}{https://github.com/open-compass/VLMEvalKit}}.

\paragraph{Model Training Framework} In order to optimize training efficiency, we employ an in-house training framework developed by WeChat. This framework improves the data loading pipeline by introducing a more efficient data format. To further enhance training efficiency, we concatenate multiple samples to achieve a maximum sequence length of 4096, which is subsequently fed into the model. This method is consistent with the standard practices used in training large language models. Furthermore, our framework supports various types of model parallelism, such as tensor parallel\cite{korthikanti2023reducing} and pipeline parallel, as well as data parallelism. Before embarking on a comprehensive exploration, it is essential to train a baseline model, for instance, LLaVA-1.5\cite{liu2023improved}, to validate the accuracy of our training framework. We maintain consistency with LLaVA-1.5 in terms of all datasets, including pre-training and supervised fine-tuning, as well as hyper-parameters. Additionally, we select several language models, such as Vicuna\cite{vicuna2023}, Qwen-1.5\cite{bai2023qwen}, and Nous-Hermes-2-Yi\cite{young2024yi}, to ensure our conclusions are more generalizable and convincing. Just as shown in Table~\ref{tab:baseline}, the model of our implementation is comparable to that of the official implementation. When the language model is replaced, models obtained by our training frameworks can also abtain reasonable results. The official implementation results are primarily sourced from the OpenCompass leaderboard\footnote{\href{https://rank.opencompass.org.cn/leaderboard-multimodal}{https://rank.opencompass.org.cn/leaderboard-multimodal}}\cite{2023opencompass}, with the exception of the ScienceQA results, which are referenced from the original paper.

\input{table/table_1}

\paragraph{Improved Pre-training Settings} The original LLaVA fixes the vision encoder, focusing solely on the MLP adapter's pre-training to enhance efficiency. However, an increasing number of studies\cite{chen2023sharegpt4v, internlmxcomposer2, chen2024far} suggest that jointly training the vision encoder and the adapter can be advantageous. This approach allows for the adjustment of the feature distribution to the generation task, thereby enhancing the vision encoder's feature extraction capabilities. In our work, we also unfreeze the vision encoder and assign different learning rates to the vision encoder and the MLP adapter. As demonstrated in Table~\ref{tab:improve_pretraining_setting}, this configuration improves LLaVA's performance non-trivially.

\input{table/table_2}

\subsection{Scaling Up the Pre-training Data}

\input{figures/data_scale}

Rather than solely relying on the 585K data from LLaVA for pre-training, a growing body of research is incorporating significantly larger datasets during this phase. For instance, Qwen-VL\cite{bai2023qwen} and InternVL\cite{chen2023internvl, chen2024far} utilize web-scale pre-training datasets such as LAION-5B\cite{schuhmann2022laionb} and COYO-700M\cite{kakaobrain2022coyo-700m}. While these studies have shown promising results, a more thorough analysis is required to fully comprehend data efficiency during the vision-language pre-training stage. 
In this section, we delve further into this topic by training a model on a subset of LAION-5B, ranging from 1M to 100M. Additionally, we investigate the relationship between the model size and the scale of the pre-training dataset. To strengthen our conclusions, we also include three types of large language models, namely Vicuna\cite{vicuna2023}, Qwen-1.5\cite{bai2023qwen}, and Hous-Hermes-2-Yi\cite{young2024yi}, in our experiments. The detailed results are depicted in Figure~\ref{fig:data-scale}.

\vspace{1em}

\noindent Three main insights can be gleaned from the trends in Figure~\ref{fig:data-scale}: 1) The pre-training datasets currently used in vision-language pre-training are quite inefficient. As we scale up the pre-training dataset from 1M to 100M, we only observe marginal improvements across the three benchmarks, and performance even deteriorates as we increase the dataset size beyond 50M. For example, the performance of Qwen1.5-7B on SEED-Bench drops 3.3 points as we scale the size of pre-training data from 20M to 100M. 2) Scaling up the size of the LLM can yield substantial improvements. For instance, Vicuna-13B outperforms Vicuna-7B by 3.0 when 1M data is used for pre-training. 3) The performance trend for different model sizes is almost consistent across different pre-training data scales. This observation contrasts with the trend in large language models, which suggests that larger models using more data can obtain further improvement.

\noindent \textit{Based on the observed experimental phenomena, it is evident that simply scaling up the size of the pre-training dataset does not effectively enhance the performance of vision-language models. A more promising approach lies in enhancing the quality and diversity of the data, as suggested by studies on LLM \cite{touvron2023llama, cai2024internlm2}, which emphasizes the importance of utilizing more than just generic image-caption pairs.}

\subsection{Instruction Dataset Selection}

Based on the taxonomy of SFT data in InternVL-1.5\cite{chen2024far}, we have added a new category related to screenshots, resulting in a total of 12 categories. For the datasets in each category, we still use the datasets given in InternVL-1.5 as the base version, and then introduce datasets used in other works, such as DeepSeek-VL\cite{lu2024deepseek}. Before conducting detailed ablation experiments on the datasets, we conduct a comprehensive study of the selected datasets and eliminated some of the lower quality ones. At the same time, we find that some datasets in InternVL-1.5 are not accurately classified. For example, ALLaVA\cite{chen2024allava} contains both picture caption data and conversation data, but InternVL-1.5 completely places it in the conversation category. Through filtering and reclassification of the datasets, we obtain a base version of the datasets before further selection (Table~\ref{tab:dataset}). 

Given that LLaVA-1.5-665K is currently the most extensively utilized dataset for visual instruction tuning, we select it as the starting point for our investigation. Moreover, as demonstrated by previous studies, substituting the \textit{Detailed Description} data in LLaVA-1.5-665K with data from ShareGPT4V can yield further enhancements. Consequently, we opt for the improved version of LLaVA-1.5-665K as our ultimate starting point. The \textbf{Individual Select} strategy then selects the most effective datasets from Figure~\ref{tab:dataset} and incorporates them into LLaVA-1.5-665K. The workflow of the \textbf{Individual Select} strategy are as follows:

\begin{enumerate}[label={\bf {{(\arabic*)}}},leftmargin=*,topsep=0.5ex,itemsep=-0.5ex,partopsep=0.75ex,parsep=0.75ex,partopsep=0pt,wide, labelwidth=0pt,labelindent=0pt]
    \item For each dataset (candidate) from a category in Table~\ref{tab:dataset}, we incorporate it into the baseline dataset and fine-tune the model on this newly constructed dataset.
    \item If the model's performance surpasses or is comparable to that achieved when trained on the baseline dataset, we include the candidate dataset in the candidate pool. If not, we discard it. Ultimately, we concatenate all the datasets from the candidate pool and integrate them into the baseline dataset to establish a new baseline dataset.
    \item We then employ this new baseline dataset, iterate through all the subsequent categories, and repeat Steps 1 and 2 above.
\end{enumerate}

In Step 2, as outlined above, we aim to avoid overfitting the four benchmark datasets. Therefore, if the model trained on the newly constructed dataset yields performance that is merely comparable to, and does not exceed, that of the model trained on the baseline dataset, we still include it in the candidate pool. Furthermore, we have observed that if two datasets individually contribute to improvements, their combination can lead to even further enhancements. Figure~\ref{fig:main} shows the details of \textbf{Individual Select} and the final datasets we choose to be added to the improved version of LLaVA-1.5-665K. he final selection of datasets we use comprises nine categories, reducing the original number from 37 to 17. 
\input{figures/main}

\input{table/table_3}

%% file: table/table_1.tex
\begin{table}[!t]
\centering
\tabcolsep 4pt
\scalebox{0.75}{
\begin{tabular}{lllll}
LLM & MME\cite{Fu2023MMEAC} & MMB-dev\cite{liu2023mmbench} & SQA$^\text{I}$\cite{lu2022learn} & SEED$^\text{I}$\cite{li2023seed} \\
\midrule 
\rowcolor{gray!15}
Vicuna-7B\cite{vicuna2023} & 1808.4 & 65.2 & 66.8 & 65.8 \\
\rowcolor{green!30}
Vicuna-7B\cite{vicuna2023} & 1772.2 \less{(-36.2)} & 64.1 \less{(-1.1)} & 70.0 \more{(+3.2)} & 65.1 \less{(-0.7)} \\[2pt]
\hdashline
Yi-6B\cite{young2024yi} & 1772.8 & 70.4 & 73.5 & 68.6 \\
Yi-34B\cite{young2024yi} & 1840.3 & 74.8 & 75.2 & 72.0 \\[2pt]
\hdashline
Qwen-1.5-7B\cite{bai2023qwen} & 1657.4 & 67.4 & 67.0 & 66.6 \\
Qwen-1.5-14B\cite{bai2023qwen} & 1801.7 & 70.6 & 71.6 & 68.2 \\
\end{tabular}
}
\caption{\textbf{Comparison with the official implementation of LLaVA-1.5.} MMB-dev: the \textit{dev} set of MMBench. SQA$^\text{I}$\cite{lu2022learn}: the image split of ScienceQA. SEED$^\text{I}$\cite{li2023seed}: the image split of Seed-Bench. The gray line illustrates the model's performance as per the official implementation, while the green line represents the model's performance achieved through our training framework. }
\vspace{-0.1cm}
\label{tab:baseline}
\end{table}

%% file: table/table_2.tex
\begin{table}[!t]
\centering
\tabcolsep 4pt
\scalebox{0.8}{
\begin{tabular}{ccllll}
lr$^\text{v}$ & lr$^\text{a}$ & MME & MMB-dev & SQA$^\text{I}$ & SEED$^\text{I}$ \\
\midrule 
\rowcolor{gray!15}
N/A & 1e-3 & 1772.2 & 64.1 & 70.0 & 65.1 \\
2e-5 & 2e-4 & 1744.0.0 \less{(-32.2)} & 65.5 \more{(+1.1)} & 70.2 \more{(+0.2)} & 66.3\more{(+1.2)} \\
\end{tabular}
}
\caption{\textbf{Improved pre-training settings.} Unfreezing the vision encoder is beneficial to improve the performance of LLaVA. lr$^\text{v}$: learning rate for vision encoder. lr$^\text{a}$: learning rate for the MLP adapter. N/A: fix the vision encoder.}
\vspace{-0.1cm}
\label{tab:improve_pretraining_setting}
\end{table}

%% file: figures/data_scale.tex
\begin{figure*}[!ht]
\vspace{-0.6em}
\centering
\includegraphics[width=1.0\linewidth]{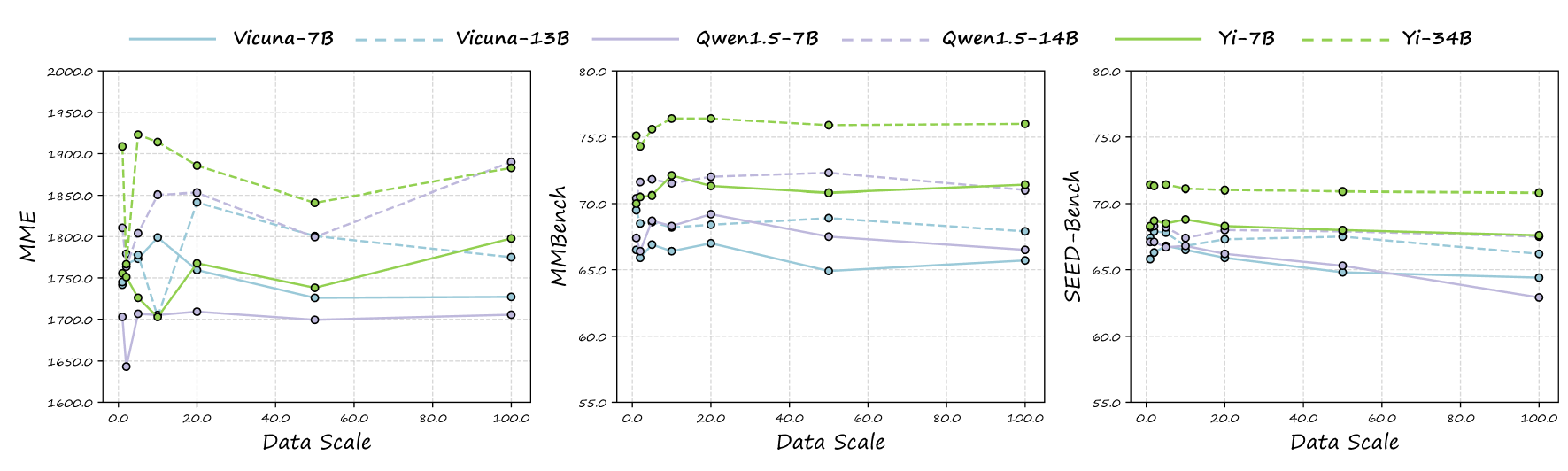}
\caption{\textbf{Pre-training data scaling law.} We investigated this phenomenon in large vision-language models using three different types of mainstream Large Language Models (LLMs). As we increased the size of the pre-training dataset from 1 million to 100 million samples, the model's performance remained nearly consistent, with some instances of degradation observed.}
\label{fig:data-scale}
\end{figure*}

%% file: figures/main.tex
\begin{figure*}[!htbp]
\centering
\includegraphics[width=\linewidth,scale=1.00]{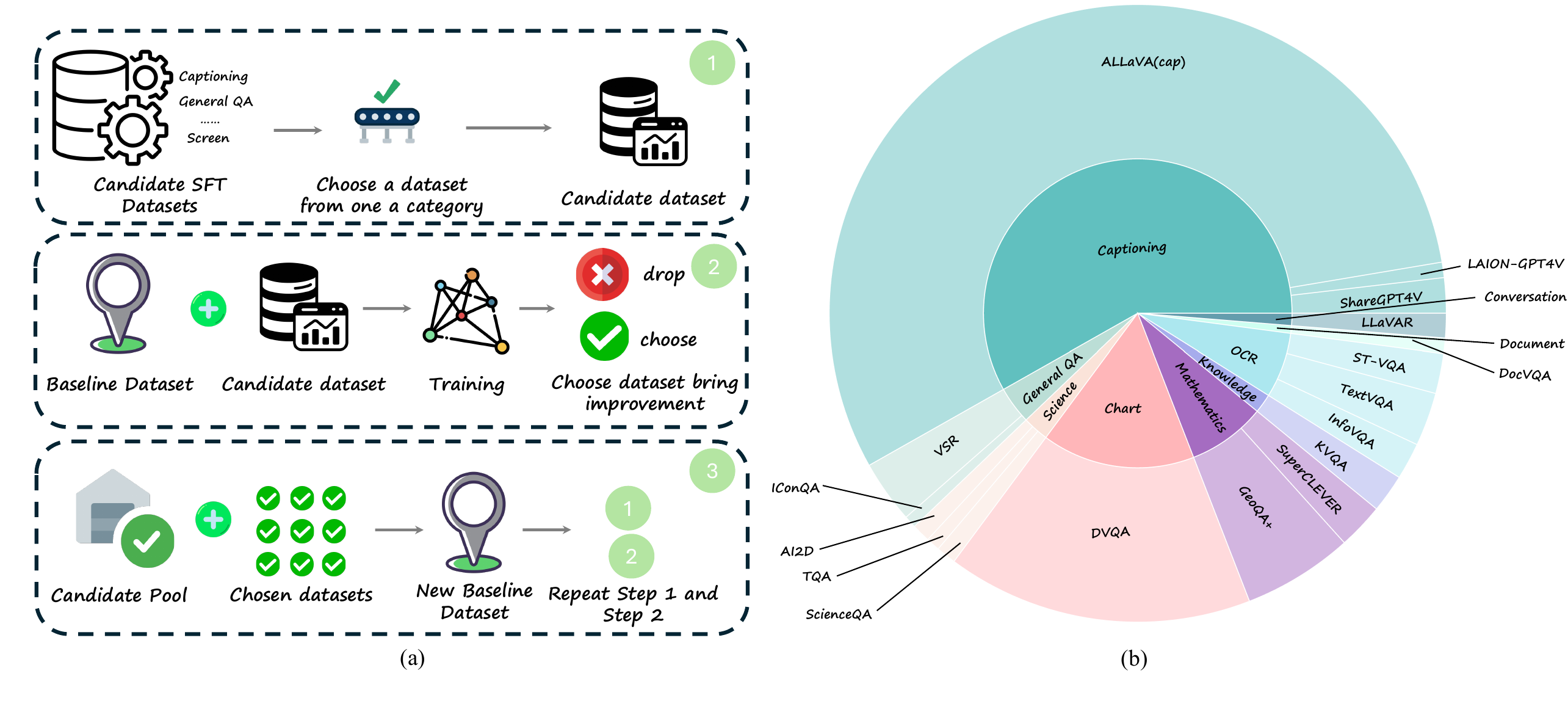}
\caption{\textbf{The workflow of Individual Select (a) and the SFT datasets we finally choose to be added to the baseline dataset.}}
\label{fig:main}
\end{figure*}

%% file: table/table_3.tex
\begin{table}[t]\scriptsize
\renewcommand{\arraystretch}{1.2}
\begin{subtable}{0.47\textwidth}
    \setlength\tabcolsep{6.4pt}
    \begin{tabular}{l|l}
Task & Dataset \\
\hline
\rowcolor{green!30}
                              & ShareGPT4V~\cite{chen2023sharegpt4v}, 
                                LAION-GPT4V, TextOCR-GPT4V~\cite{textocr-gpt4v} \\
\rowcolor{green!30}
\multirow{-2}{*}{Captioning}  & SVIT(cap)~\cite{zhao2023svit},           
                                ALLaVA(cap)~\cite{chen2024allava}, 
                                LVIS-Instruct4V(cap)~\cite{wang2023see}\\
                    
\rowcolor{gray!15}
\multirow{-1}{*}{General QA}  & VSR~\cite{Liu2022VisualSR}, IConQA~\cite{lu2021iconqa}              \\
\rowcolor{green!30}
\multirow{-1}{*}{Science}     & AI2D~\cite{kembhavi2016diagram}, ScienceQA~\cite{lu2022learn}, TQA~\cite{Kembhavi2017AreYS}      \\
\rowcolor{gray!15}
                              & ChartQA~\cite{masry-etal-2022-chartqa}, MMC-Inst\cite{liu2023mmc}, DVQA (en)\cite{kafle2018dvqa},     \\
\rowcolor{gray!15}
\multirow{-2}{*}{Chart}       & PlotQA~\cite{Methani_2020_WACV}, UReader~\cite{ye2023ureader}                     \\
\rowcolor{green!30}
                              & GeoQA+~\cite{cao-xiao-2022-augmented}, TabMWP (en)~\cite{lu2023dynamic}      \\
\rowcolor{green!30}
\multirow{-2}{*}{Mathematics} & CLEVR-Math/Super~\cite{li2023super, lindstrom2022clevr}, Geometry3K\cite{lu2021inter} \\           
\rowcolor{gray!15}
\multirow{-1}{*}{Knowledge}   & KVQA~\cite{shahMYP19} \\                          
\rowcolor{green!30}
                              & InfoVQA~\cite{mathew2022infographicvqa}, TextVQA~\cite{singh2019towards}, ArT~\cite{chng2019icdar2019}\\ 
\rowcolor{green!30}
\multirow{-2}{*}{OCR}         & SynthDoG~\cite{kim2022donut}, ST- 
                              VQA~\cite{biten2019scene} \\
\rowcolor{gray!15}
Document                      & DocVQA~\cite{mathew2021docvqa}, \\
\rowcolor{green!30}
Grounding                     & RefCOCO+/g~\cite{yu2016modeling, mao2016generation}                         \\
\rowcolor{gray!15}
                              & ALLaVA(conv)~\cite{chen2024allava}, LVIS-Instruct4V(conv)~\cite{wang2023see} \\
\rowcolor{gray!15}
\multirow{-3}{*}{Conversation} & SVIT(conv)~\cite{zhao2023svit}, LLaVAR~\cite{zhang2023llavar}, VisualDialog~\cite{das2017visual}   \\
\rowcolor{green!30}
\multirow{-1}{*}{Text-only}  & OpenHermes2.5~\cite{OpenHermes-2.5}, Alpaca-GPT4~\cite{alpaca}, LIMA~\cite{zhou2024lima}                        \\
\rowcolor{gray!15}
\multirow{-1}{*}{Screen}  & ScreenQA~\cite{baechler2024screenai} \\
\end{tabular}
\label{tab:finetuning}
\end{subtable}
\caption{\textbf{Base version of SFT datasets to be ablated.} We utilize the \textbf{Individual Select} strategy to incrementally select the most effective datasets from the aforementioned ones, and incorporate them into LLaVA-1.5-665K to enhance the richness of the SFT dataset. cap: the caption split. conv: the conversation split.
}
\label{tab:dataset}
\end{table}

%% file: sec/4-exp.tex
\section{Experiments}
\label{sec:exp}

\noindent \paragraph{Experiment Setting} Apart from the modifications introduced in Section~\ref{sec:method}, all other settings remain consistent with those of LLaVA-1.5. Specifically, we employ OpenAI's CLIP-Large-336px\cite{radford2021learning} as the vision encoder. The learning rate is linearly warmed up during the initial 3\% of iterations, after which a cosine decay learning rate strategy is implemented.

\subsection{Comparison with Other Models}

In this section, we provide a thorough comparison of our model, which utilizes techniques from Section~\ref{sec:method}, with other state-of-the-art (SOTA) models across seven benchmarks, namely MMBench\cite{liu2023mmbench}, MME\cite{Fu2023MMEAC}, MathVista\cite{lu2023mathvista}, HallusionBench\cite{liu2023hallusionbench}, SEEDBench\cite{li2023seed}, and LLaVABench\cite{liu2024visual}. We have chosen two models for this comparison: Vicuna-1.5-7B/13B. For both models, we select the best pre-trained versions as identified by the ablation study illustrated in Figure~\ref{fig:data-scale}. The detailed results are presented in Table~\ref{tab:comparison}. As the table indicates, when comparing our model with other models using the same LLM, such as Vicuna-7B/13B, our model outperforms the others by a significant margin overall. Despite being directly based on LLaVA-1.5, our model even surpasses LLaVA-Next, which introduces new strategies like the use of high-resolution images. This substantial performance improvement further underscores the importance of exploring the composition of SFT datasets. 


\input{table/table_4}

\input{table/table_5}

\subsection{Ablation Studies}

\paragraph{Meticulously choose the SFT datasets is important.} In this section, we compare the models fine-tuned on all these datasets from Table~\ref{tab:dataset} and datasets obtained by \textbf{Individual Select}. Just as shown, indiscriminately fine-tuning the model on the all datasets will not bring improvements, or even degrade the performance. Another drawbacks of fine-tuning the model on all the datasets is the heavy computation overhead, since the size of these datasets is about six times of ours.

\paragraph{Consistent improvements.} The paragraph, as depicted in Figure~\ref{fig:sft_improve}, demonstrates a consistent overall improvement during the dataset selection process as more datasets from each category are incorporated. We also observe a nearly linear improvement trend for each benchmark, with the exception of MME. However, the performance trend of MME remains relatively stable after the inclusion of datasets from the \textit{Caption} category. Notably, there is a significant improvement in ScienceQA following the introduction of datasets from the \textit{Science} category. This can be attributed to the addition of the \textit{train split} of ScienceQA. This observation underscores the potential for enhancing model performance on a specific task by introducing a dataset with a similar distribution to that task. However, it is crucial to maintain a balance with other datasets to prevent a decline in the model's general capabilities.
\input{figures/sft_improvement}

%% file: table/table_4.tex
\begin{table*}[!ht]
\centering
\tabcolsep 7pt
\scalebox{0.75}{
\begin{tabular}{lllccccccc}
Method  &  LLM  &  Vision  &  MMBench (en)  &  MMBench (cn)  &  MME  &  MathVista  &  HallusionBench & SEED$^\text{I}$ & LLaVABench \\
\midrule
LLaVA-1.5\cite{liu2023improved} & Vicuna-7B\cite{vicuna2023} & CL\cite{radford2021learning} & 64.3 & 58.3 & \underline{1808.4} & 25.6 & 27.6 & 66.1 & 65.4 \\
LLaVA-Next\cite{liu2024llavanext} & Vicuna-7B\cite{vicuna2023} & CL\cite{radford2021learning} & \underline{69.2} & \underline{62.3} & 1769.1 & 31.5 & 27.6 & \underline{69.6} & \underline{72.7} \\
CogVLM\cite{wang2023cogvlm} & Vicuna-7B\cite{vicuna2023} & E2CL-E\cite{sun2023eva} & 65.8 & 55.9 & 1736.6 & \underline{35.0} & \textbf{35.4} & 68.8 & \textbf{73.9} \\
Ours & Vicuna-7B\cite{vicuna2023} & CL\cite{radford2021learning} & \textbf{72.6} & \textbf{65.8} & \textbf{1818.7} & \textbf{42.6} & \underline{31.9} & \textbf{69.9} & 62.7 \\
\hdashline
LLaVA-1.5\cite{liu2023improved} & Vicuna-13B\cite{vicuna2023} & CL\cite{radford2021learning} & 67.7 & 63.6 & 1780.8 & 27.7 & 24.5 & 68.2 & 66.1 \\
LLaVA-Next\cite{liu2024llavanext} & Vicuna-13B\cite{vicuna2023} & CL\cite{radford2021learning} & 68.8 & 61.9 & 1745.6 & \underline{34.1} & \textbf{31.8} &70.1  & \textbf{73.9} \\
ShareGPT4V\cite{chen2023sharegpt4v} & Vicuna-13B\cite{vicuna2023} & CL\cite{radford2021learning} & \underline{69.8} & \underline{65.1} & \underline{1853.1} & 29.3 & 28.4 & \textbf{70.6} & \underline{69.1} \\
Ours & Vicuna-13B\cite{vicuna2023} & CL\cite{radford2021learning} & \textbf{74.9} & \textbf{69.8} & \textbf{1879.8} & \textbf{39.1} & \underline{31.4} & \underline{70.3} & 65.2 \\
\hdashline

\end{tabular}
}
\caption{\textbf{Comparison with other methods across benchmarks for instruction-following LMMs.} We select the evaluation metric for each method based on its original paper, if available. If not, we use the metric provided by the leaderboard of VLMEvalKit. Vision: vision encoder. CL: OpenAI CLIP-L-336px. E2CL-E: EVA2-CLIP-E.}
\label{tab:comparison}
\end{table*}

%% file: table/table_5.tex
\begin{table}[h]
\centering
\tabcolsep 4pt
\scalebox{0.8}{
\begin{tabular}{llllll}
Dataset & LLM & MME & MMB-dev & SQA$^\text{I}$ & SEED$^\text{I}$ \\
\midrule 
All & Vicuna-7B & 1790.0 & 70.5 & 80.1 & 70.0 \\
Ours & Vicuna-7B & 1818.7 \more{(+28.7)} & 73.0 \more{(+2.5)} & 81.6 \more{(+1.5)} & 69.9 \less{(-0.1)} \\
\end{tabular}
}
\caption{\textbf{Model performance comparison between fine-tuned on all datasets and datasets we select.} All: all datasets from Table~\ref{tab:dataset}.}
\vspace{-0.1cm}
\label{tab:all}
\end{table}

%% file: figures/sft_improvement.tex
\begin{figure}[!htbp]
\centering
\includegraphics[width=\linewidth,scale=1.00]{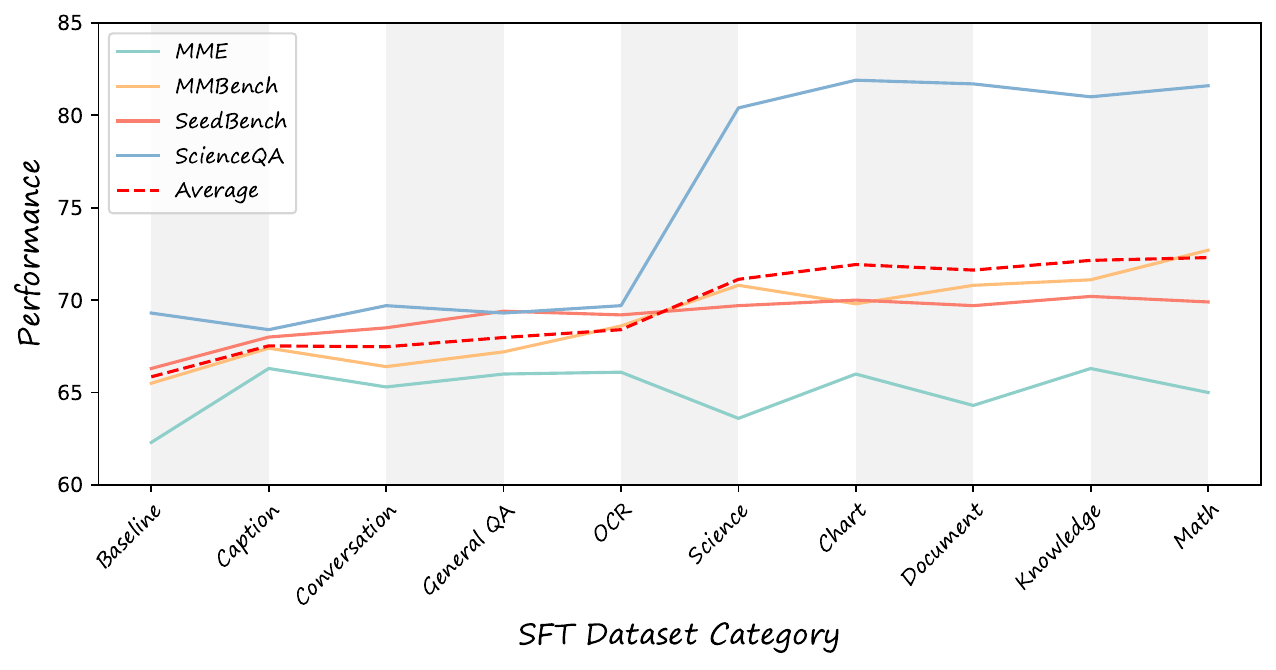}
\caption{\textbf{We observe consistent improvements as we incorporate these selected datasets from each category.} The original MME scores are mapped to a range of 0 to 100. Average: the average score of the four metrics.}
\label{fig:sft_improve}
\end{figure}

%% file: sec/5-discussion.tex
\section{Discussion, Limitation and Future Works}
\label{tab:disc}

\paragraph{Discussion} As discussed in Section~\ref{sec:method}, we find that simply increasing the size of the pre-training dataset does not consistently yield improvements. We propose two possible reasons for this: i) The quality of the pre-training dataset may be suboptimal. LAION-5B, which is crawled from the Internet and only subjected to basic data filtering processes such as image-text similarity filtering, may contain a significant amount of noise, including grammatical errors in text and incorrect punctuation. ii) The vision encoder has already been pre-trained on a dataset with a distribution similar to that used in the vision-language alignment pre-training. Given that the vision encoder is not frozen, the vision-language pre-training focuses on vision-language alignment and the injection of new knowledge into the vision encoder. This new knowledge injection involves instilling abilities that were not learned during the vision encoder pre-training. Therefore, using more data with a distribution similar to that used for vision encoder pre-training does not result in substantial improvements during the vision-language pre-training stage.

\paragraph{Limitation} In our exploration of SFT datasets, we try to prevent overfitting. The model, fine-tuned on our dataset, also performs well on other benchmarks as shown in Table~\ref{tab:comparison}. However, the performance gap on other benchmarks is smaller than that on the three benchmarks used in the ablation study, suggesting that more appropriate benchmarks for ablation should be selected. Currently, our investigation of the SFT dataset is primarily conducted at the level of individual datasets. While this approach has led to significant improvements in model performance, it is relatively rudimentary. Furthermore, we have not yet explored aspects related to the quality and distribution of the SFT dataset in our current work.

\paragraph{Future Works} 

The primary objective of this paper is not to introduce a state-of-the-art (SoTA) model that performs competitively across a variety of benchmarks. Instead, our focus is on investigating some aspects that have been previously overlooked. While the model we have developed performs satisfactorily on a series of benchmarks, there is room for further refinement in terms of user experience. In future work, we will concentrate on three main areas: i) Dataset quality: This includes both pre-training and SFT datasets. We plan to undertake a series of explorations on how to clean existing datasets and generate more effective ones. ii) Knowledge injection: We aim to move beyond solely relying on general image-text caption datasets. Our intention is to incorporate datasets with different distributions during pre-training, such as Optical Character Recognition (OCR). iii) Incorporation of recent techniques in Vision-Language Models (VLM): This includes the use of high-resolution images.

%% file: sec/6-conclusion.tex
\section{Conclusion}

In this study, we delve into critical yet previously neglected aspects of vision-language models, such as the scaling law during pre-training and the selection of the most effective dataset for instruction fine-tuning. We consider three types of Language Learning Models (LLMs) and assemble seven splits of pre-training datasets, with the total count ranging from 1M to 100M. Our extensive experiments reveal that simply increasing the size of the pre-training dataset does not necessarily yield significant improvements and may even degrade the model's performance. Furthermore, we introduce a strategy, termed \textbf{Individual Select}, to identify the most effective datasets from a vast pool of publicly available candidates. This approach leads us to an effective composition of instruction tuning (SFT) datasets. Models fine-tuned on these datasets demonstrate substantial improvements over the baseline and outperform models that are indiscriminately fine-tuned on all SFT datasets. This underscores the need for a more thoughtful composition of SFT datasets.